# cito: An R package for training neural networks using torch

Christian Amesöder[1,2], Florian Hartig[1], Maximilian Pichler[1,*]

[1] Theoretical Ecology, University of Regensburg, Universitätsstraße 31, 93053 Regensburg, Germany

[2] Information Systems, University of Regensburg, Universitätsstraße 31, 93053 Regensburg, Germany

* corresponding author, maximilian.pichler@biologie.uni-regensburg.de

## Abstract

Deep Neural Networks (DNN) have become a central method in ecology. Most current deep learning (DL) applications rely on one of the major deep learning frameworks, in particular Torch or TensorFlow, to build and train DNN. Using these frameworks, however, requires substantially more experience and time than typical regression functions in the R environment. Here, we present 'cito', a user-friendly R package for DL that allows specifying DNNs in the familiar formula syntax used by many R packages. To fit the models, 'cito' uses 'torch', taking advantage of the numerically optimized torch library, including the ability to switch between training models on the CPU or the graphics processing unit (GPU) (which allows to efficiently train large DNN). Moreover, 'cito' includes many user-friendly functions for model plotting and analysis, including optional confidence intervals (CIs) based on bootstraps for predictions and explainable AI (xAI) metrics for effect sizes and variable importance with CIs and p-values. To showcase a typical analysis pipeline using 'cito', including its built-in xAI features to explore the trained DNN, we build a species distribution model of the African elephant. We hope that by providing a user-friendly R framework to specify, deploy and interpret DNN, 'cito' will make this interesting model class more accessible to ecological data analysis. A stable version of 'cito' can be installed from the comprehensive R archive network (CRAN).

# Introduction

Deep neural networks (DNN) are increasingly used in ecology and evolution for regression and classification tasks such as species distribution models, image classification or sound analysis (Christin et al., 2019; Joseph, 2020; Pichler and Hartig, 2023a; Strydom et al., 2021). State-of-the-art DNN are almost exclusively implemented and trained in specialized deep learning (DL) frameworks such as PyTorch or Tensorflow (Abadi et al., 2016; Paszke et al., 2019). These frameworks, most of which are implemented in Python, provide flexible and performant functions and classes that allow users to implement and train complex DL architectures, such as large language models (e.g., GPT-3 (Brown et al., 2020), RoBERTA (Liu et al., 2019)) or complex object detection models (e.g., Mask R-CNN (He et al., 2017), DeepVit (Zhou et al., 2021)). Their high level of flexibility is appealing to "power users", but the complexity of these frameworks can be prohibitive or at least repelling for scientists with limited knowledge in the field that merely want to use neural networks in standard applications.

As a response to this problem, several simplified frontends for the major DL frameworks have been developed. Many of those are also available in R, the language used by most ecologists for practical data analysis. Well-known examples are 'Keras' for TensorFlow and luz for 'torch' (Allaire and Chollet, 2022; Falbel, 2022). However, while these frontends indeed simplify the model building process considerably, their general structure and syntax still resembles those of the major Python frameworks rather than those of popular R packages for regression or classification tasks that specify models using the formula syntax such as 'ranger', for training random forests, or 'lme4', for training mixed-effect models (Bates et al., 2015; Wright and Ziegler, 2017). Moreover, DL frontends such as 'Keras' or 'luz' mainly concentrate on model fitting and include only a very limited set of plots and convenience functions which are common to most R packages. As a result, working with these frontends still requires a considerable amount of training for users that are so far only familiar with standard R packages. Especially because users have to choose or program code for downstream tasks such as bootstrapping, plots or explainable AI (xAI) metrics by hand.

Besides the mentioned frontends to the major DL frameworks, some specialized R packages for training DNN exist that more closely adhere to the syntax used in most popular R packages, in particular the formula syntax to specify the model structure. However, those packages often lack crucial functionalities, and most of them do not make use of state-of-the-art DL frameworks for model fitting. This limits their use for large DNN because of their numerical inefficiency or their inability to train the models on GPUs. Established R packages such as 'nnet' or 'neuralnet' do not support modern DL techniques, such as different regularization techniques (e.g. dropout) to control the bias-variance tradeoff (Fritsch et al., 2019; Venables and Ripley, 2002) or modern training techniques such as early stopping or learning rate schedulers that help to achieve convergence. The 'h2o' package comes with its own Java backend, and while it allows specifying models with the standard formula syntax, its use in R is cumbersome due to its inability to work with default R objects (LeDell et al., 2022). The 'brulee' R package (Kuhn and Falbel, 2022), which uses 'torch' to train the DNNs specified in standard R syntax, is very similar to the package presented here, but still lacks some critical features (see section 'Performance analysis and validation').

Here, we present 'cito', an R package for training fully-connected neural networks using the standard R formula syntax for model specification. Based on the 'torch' DL framework, 'cito' allows flexible specifying of fully-connected neural networks architectures, supports many modern DL techniques (e.g. dropout and elastic net regularization, learning rate schedulers), can take advantage of CPU and GPU hardware for parallelization, and, despite its simple user interface, optionally offers a high degree of customization such as user-defined loss functions. Moreover, 'cito' supports many downstream functionalities, such as the possibility to continue the training of existing DNN with modified training parameters for fine-tuning, or the application of explainable AI (xAI) methods to interpret the trained models. As such, 'cito' provides a user-friendly but nevertheless complete analysis pipeline for building neural networks in R.

In the remainder of the paper, we introduce the design principles of 'cito' in more detail, show validation and performance analysis, and showcase the application of cito using the example of a species distribution model of the African elephant.

# Design of the cito package

## Torch backend

'cito' uses 'torch', a variant of PyTorch, as its backend to represent and train the specified neural networks. Until recently, R users who wanted to use PyTorch and Tensorflow had to call their Python bindings through the 'reticulate' package. R packages that relied on this pipeline were thus dependent on appropriate Python installations (e.g. Pichler and Hartig, 2021), which often created dependency issues. This issue got solved with the release of 'torch', a native implementation of the torch libraries with an R frontend (Falbel and Luraschi, 2022).

## Building and training neural networks in cito

With 'torch', R users can essentially use PyTorch natively in R, which solves dependency issues, but not the problem that specifying a DNN with 'torch' is complex.

'cito' addresses this problem by providing one simple command, *dnn()*, which combines everything needed to build and train a fully-connected neural network in one line of code (see package *vignette('A-Introduction_to_cito')* for more details). The *dnn()* function includes options to modify the network architecture, the training process and the monitoring (e.g. by visualization) of the training and validation loss (

Table *1*), including a baseline loss (based on intercept-only models) that helps to diagnose convergence problems due to inappropriately chosen training hyperparameters (e.g., learning rate and epochs).

The *dnn()* function returns an S3 object that can be used, for example, with the *continue_training()* function to continue training for additional epochs (iterations) with the same or modified training hyperparameters or data. Moreover, many standard R functions such as *summary()*, *predict()* or *residuals()* are implemented for the trained models, and additional specialized explainable xAI functions are available for interpreting the fitted networks. More details on these and other functions are available in the R package vignettes that come with the cito package.

The lack of uncertainties (standard errors) is an often-raised concern for DNN. In 'cito', we provide an option to automatically calculate confidence intervals for all outputs (including xAI metrics and predictions) using bootstrapping. As bootstrapping can be computationally expensive, the default for this option is set to false. Bootstrapping can be enabled in the *dnn()* function setting, e.g., *dnn(… ,bootstrap = 50)*. Bootstrap standard errors are then automatically propagated through all downstream methods and are also used to generate p-values wherever obvious null hypotheses exist. We recommend starting without bootstrapping to optimize the training procedure (Fig. 2) and to then enable the bootstrap for the final model after the training pipeline has been finalized.

**Table 1:** Hyperparameters for fully-connected neural networks and their default values in 'cito'. Defaults for all parameters are set to sensible values; however, some parameters typically need to be tuned. Detailed guidance on this is provided in the help file of the *dnn*() function or in the cito R package vignette 'Training neural networks'.

**Architecture**

| Name | Explanation | Default |
|---:|---|---|
| hidden | Quantity and size of hidden Layers | (50, 50) |
| activation | Activation function for hidden layers | "selu" |
| bias | Should hidden nodes have bias | TRUE |

**Training**

| Name | Explanation | Default |
|---:|---|---|
| validation | Split data into test and validation set | 0 |
| epochs | Number of training iterations | 100 |
| device | Set to "cuda" to train on GPU | "cpu" |
| plot | Visualize loss during training | TRUE |
| batchsize | Number of samples used for each training step | 32 |
| shuffle | Shuffle batches in between epochs | TRUE |
| lr | Learning rate | 0.01 |
| early_stopping | Stops training early based on validation loss | FALSE |
| Boostrap | Number of bootstrap samples | FALSE |

**Controlling bias-variance trade-off (regularization)**

| Name | Explanation | Default |
|---:|---|---|
| lambda | Strength of elastic net regularization | 0 |
| alpha | Split of L1 and L2 regularization | 0.5 |
| dropout | Dropout probability of a node | 0 |

# Performance comparison and validation of cito

After explaining the design of cito, we shortly compare its performance and functionality with other packages for implementing neural networks in R. We consider in particular 'nnet' and 'neuralnet', which each have their own backend and are not based on modern DL frameworks (Fritsch et al., 2019; Venables and Ripley, 2002), 'h2o', which possesses a much broader toolkit for training neural networks than the previous two packages (LeDell et al., 2022), and 'brulee' (Kuhn and Falbel, 2022), which, similar to cito, uses the 'torch' DL framework as a backend.

Our comparison shows that 'cito' implements more options than other packages, in particular GPU support, the possibility to continue training and custom loss functions and most importantly tools to interpret the trained DNN models (Table 2).

**Table 2:** Feature comparison of R packages used to build fully-connected neural networks

|  | 'cito' | 'brulee' | 'h2o' | 'neuralnet' | 'nnet' |
|---|---|---|---|---|---|
| Customizable network architecture | X | X | X | X |  |
| Fit a probability distribution | X |  | X |  |  |
| GPU support | X |  |  |  |  |
| Regularization | X | X | X | X | X |
| Custom loss function | X |  |  | X |  |
| Optimization of additional user-defined parameters | X |  |  |  |  |
| Continue training | X |  |  |  |  |
| Class weights for imbalanced data |  | X |  |  |  |
| Learning rate scheduler | X | X | X |  |  |
| Feature importance (xAI) | X |  | X |  |  |
| Partial dependency plots (xAI) | X |  |  |  |  |
| Accumulated local effect plots (xAI) | X |  |  |  |  |
| Uncertainty (confidence intervals and p-values for xAI metrics and predictions) | X |  |  |  |  |
| Baseline loss (to help with the convergence) | X |  |  |  |  |

Looking at computational performance, measured by the time it takes to train the networks, we find that some of the older packages, in particular 'neuralnet', perform better than the torch-based packages (including 'cito') for small networks (Figure 1). This is probably due to the smaller overhead of these more specialized packages. However, when moving to larger networks (large and especially wide networks are often beneficial for achieving low generalization errors (Belkin et al., 2019)) 'cito' can play out one of the main advantage of modern ML frameworks, which is GPU support. On the GPU, training time in cito is practically independent of the size of the network, confirming the consensus that training large networks requires GPU resources. On a CPU, 'cito' performs on par with 'brulee', the other torch-based package, but somewhat worse than 'neuralnet'. We interpret these results as showing that for a simple problem, there is still some overhead of using 'torch' as opposed to a native C implementation. Nevertheless, we would argue that the added flexibility and functionality of cito outweighs this advantage of 'neuralnet'. Moreover, our results suggest that the difference between the torch packages and 'neuralnet' lies mainly in the constant overhead needed to set up the models. For large models, their performance is roughly equal.

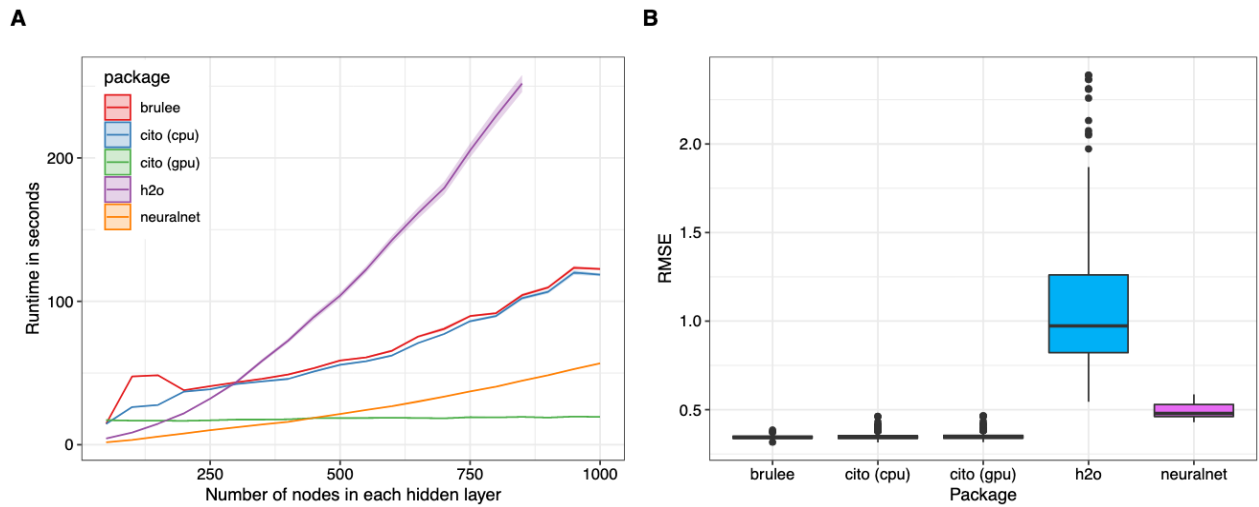

**Figure 1:** Runtime comparison of different deep learning R software packages ('brulee', 'h2o', 'neuralnet', and 'cito' (CPU and GPU)) on different network sizes on an Intel Xeon 6128 and a Nvidia RTX 2080ti. The networks consisted of five equally sized layers (50 to 1000 nodes with a step size of 50) and are trained on a simulated data set with 1000 observations. Panel (A) shows the runtime of the different packages and panel (B) shows the average root mean square error (RMSE) of the models on a holdout of size 1000 observations (RMSE was averaged over different network sizes). Each network was trained 20 times (the dataset was resampled each time).

## Workflow and case study

So far, we have mainly discussed the process of model training, which is arguably the core of any machine learning project. Now, we want to comment on the entire workflow when using 'cito' to build and interpret a predictive model. This workflow usually consists of model specification, training, and interpretation and predictions (Figure 2). To make the discussion of the workflow more accessible to the reader, we illustrate this workflow with the example (based on Ryo et al. (2021)) of building a species distribution model (SDM) for the African elephant (*Loxodonta Africana*).

SDMs are niche models that correlate environment with species occurrence data (see Elith & Leathwick, 2009). As occurrence data, we use records of African elephant presence from Ryo et al. (2021) that was based on Angelov, 2020, who compiled data from different studies available on GBIF (INaturalist Contributors, 2022a, 2022b; Jlegind, 2021; Musila et al., 2019; Navarro, 2022). Those presence-only data were supplemented by Angelov, 2020 with randomly sampled background points (pseudo-absences) to generate a presence-absence signal for the classifier. As predictors, we used all 19 bioclimatic variables from WorldClim v2 (Fourcade et al., 2018), which were centered and standardized. While it is common in statistical modelling to sample more pseudo-absences than presences, such unbalanced class numbers can be harmful for machine learning algorithms. We therefore randomly undersampled pseudo-absences to match the number of observations (another option would be to oversample presences, but in our example, this resulted in lower accuracy in interim results).

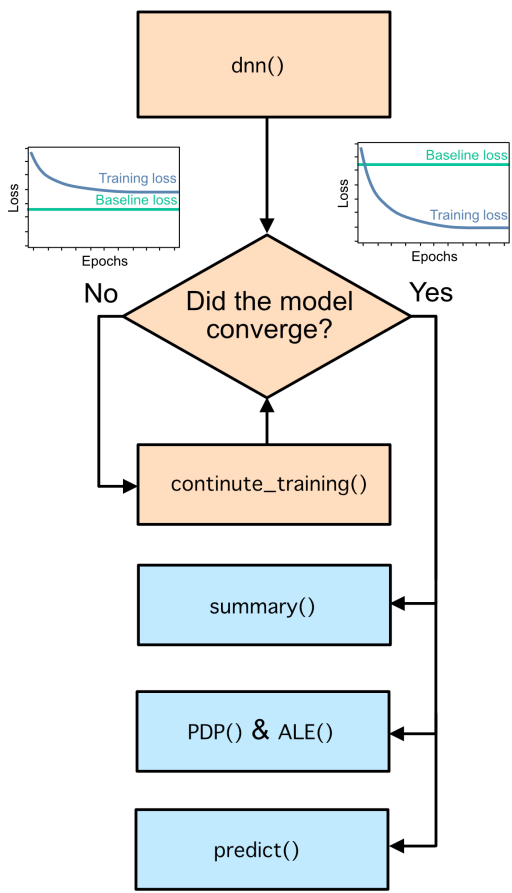

**Figure 2:** Workflow of building, training and analyzing DNN with 'cito'. Example workflow and analyses for (multi) species distribution models are available as a vignette (*vignette('C-Example_Species_distribution_modeling')*) or at https://citoverse.github.io/cito/articles

Building and training a species-distribution model based on a fully-connected neural network with three hidden layers of 50, 50 and 50 nodes and trains it for 50 epochs can be done in one line of code:

```
nn.fit <- dnn(label~., data = data,
    hidden = c(50, 50, 50), loss = "binomial",
    epochs = 50, lr = 0.1,
    batchsize = 300,
    validation = 0.1, shuffle = TRUE,
    alpha = 0.5, lambda = 0.005,
    early_stopping = 10,
    boostrap = 30)
```

During training and without bootstrapping, a plot is displayed in R that monitors the training, validation and baseline loss. This plot can be used to diagnose convergence problems, for example if the training loss does not decrease over time or does not fall below the baseline loss. In this case, it would be advisable to abort and restart the training with different hyperparameters (e.g., smaller learning rate), use a learning rate scheduler, or perform a systematic hyperparameter tuning. We provide extensive help on this topic in the documentation and in a vignette (*vignette('B-Training_neural_networks')*). Here we show an example where we restart the training with a smaller learning rate and a learning rate scheduler that automatically reduces the learning rate if the loss does not decrease in 8 continuous epochs (patience =8) to achieve a better fit:

nn.fit <- continue_training(

nn.fit, epochs = 150,

changed_params = list(lr = 0.05, lr_scheduler = config_lr_scheduler("reduce_on_plateau"), patience = 8, factor = 0.8))

The trained models can be used with a range of in-build functions. The predict() can be used to predict the occurrence probability of the elephant (Fig. 3a). The summary() function provides an overview about influential variables by calculating their importances (Fisher et al., 2019) as well as average conditional effects (which are an approximation of linear effects, see (Pichler and Hartig, 2023b)) (Fig. 4a). Partial dependency plots (PDP) and averaged local effect plots (ALE) functions can be used to display the effect of specific features on the response, in this case the occurrence probability of the elephant (Fig. 4b, c). If bootstrapping is enabled, 'cito' automatically uses the bootstrap samples to calculate confidence intervals (CI, as standard errors) for the predictions (Fig. 3a), CIs and p-values for the xAI metrics (Fig. 4a), and CIs for the PDP and ALE plots (Fig. 4b, c).

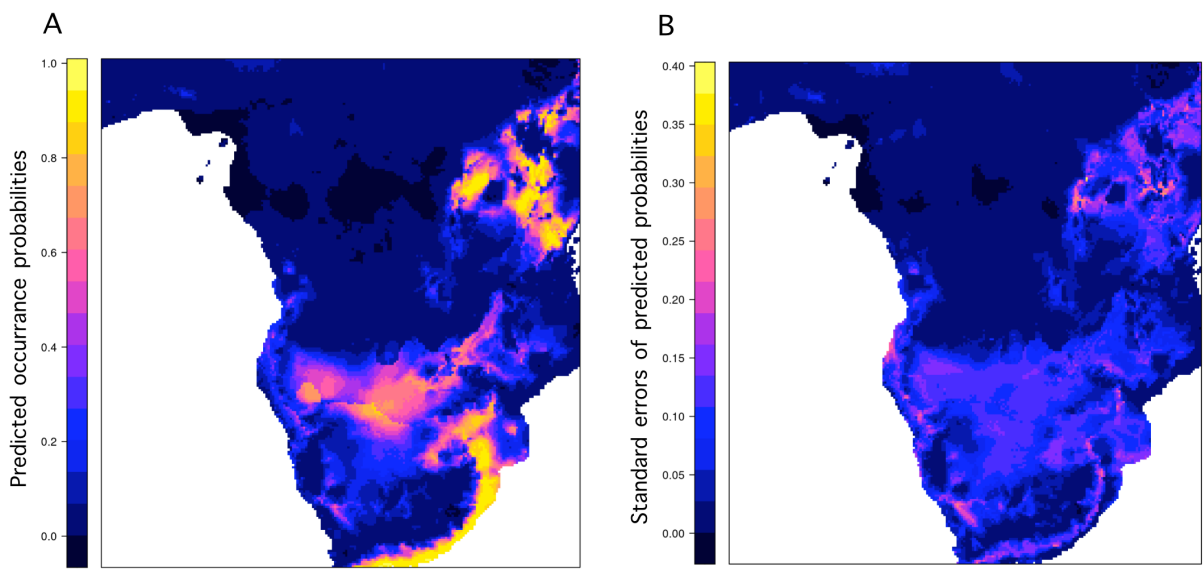

**Figure 3**: Predictions and standard errors of prediction for the African elephant from a DNN trained by cito. Panel (A) shows the predicted probability of occurrence of the African elephant. Panel (B) shows the standard error for the predicted probabilities (confidence interval).

A
```
— Feature Importance

           Importance Std.Err Z value Pr(>|z|)
bio1  → label  0.023616 0.008313    2.84   0.0045 **
bio2  → label  0.012197 0.008072    1.51   0.1308
bio3  → label  0.007474 0.008323    0.90   0.3692
bio4  → label  0.034321 0.010879    3.15   0.0016 **
bio5  → label  0.010823 0.008623    1.26   0.2095
bio6  → label  0.017014 0.007766    2.19   0.0285 *
bio7  → label  0.003234 0.003104    1.04   0.2974
bio8  → label  0.014676 0.006043    2.43   0.0152 *
bio9  → label  0.030705 0.017345    1.77   0.0767 .
bio10 → label  0.005850 0.004487    1.30   0.1923
bio11 → label  0.022184 0.008805    2.52   0.0118 *
bio12 → label  0.030492 0.014029    2.17   0.0297 *
bio13 → label  0.008422 0.007007    1.20   0.2293
bio14 → label  0.025422 0.010734    2.37   0.0179 *
bio15 → label  0.036257 0.020248    1.79   0.0733 .
bio16 → label  0.056378 0.016814    3.35   0.0008 ***
bio17 → label  0.008286 0.004852    1.71   0.0877 .
bio18 → label  0.031117 0.007020    4.43  9.3e-06 ***
bio19 → label  0.000587 0.001422    0.41   0.6795
---
Signif. codes:  0 '***' 0.001 '**' 0.01 '*' 0.05 '.' 0.1 ' ' 1

— Average Conditional Effects
              ACE Std.Err Z value Pr(>|z|)
bio1  → label  0.374   0.193    1.93  0.05300 .
bio2  → label -0.364   0.208   -1.76  0.07925 .
bio3  → label -0.154   0.292   -0.53  0.59840
bio4  → label  0.778   0.317    2.45  0.01426 *
bio5  → label  0.434   0.281    1.54  0.12285
bio6  → label  0.456   0.184    2.48  0.01322 *
bio7  → label -0.136   0.147   -0.92  0.35507
bio8  → label -0.324   0.247   -1.31  0.18882
bio9  → label -0.799   0.329   -2.43  0.01504 *
bio10 → label -0.217   0.170   -1.28  0.20168
bio11 → label -0.262   0.240   -1.09  0.27378
bio12 → label -1.094   0.312   -3.51  0.00045 ***
bio13 → label  0.584   0.289    2.02  0.04305 *
bio14 → label  0.879   0.381    2.31  0.02097 *
bio15 → label -0.707   0.334   -2.12  0.03406 *
bio16 → label -1.310   0.271   -4.83  1.4e-06 ***
bio17 → label  0.191   0.295    0.65  0.51707
bio18 → label  0.204   0.252    0.81  0.41849
bio19 → label -0.143   0.191   -0.75  0.45554
---
Signif. codes:  0 '***' 0.001 '**' 0.01 '*' 0.05 '.' 0.1 ' ' 1
```

B
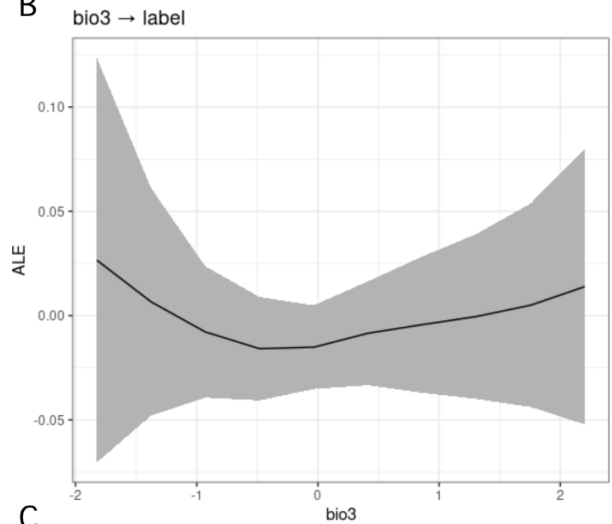

C
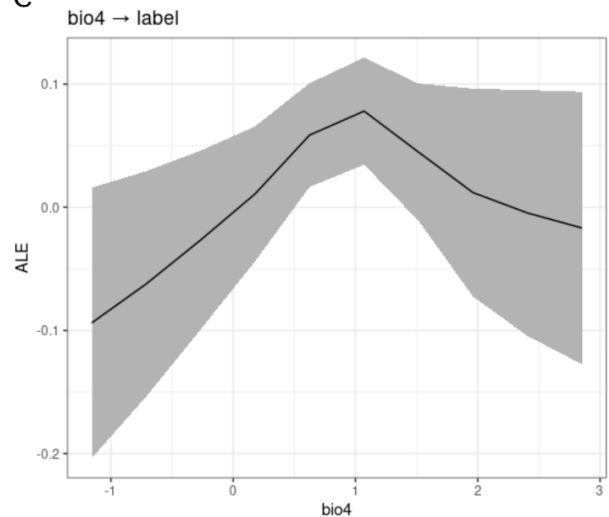

**Figure 4**: xAI metrics with bootstrap confidence intervals (+/- 1 se) from model trained by 'cito'. Panel (A) shows (permutation) feature importances and average conditional effects (approximation of linear effects) from the *summary()* output for the 19 Bioclim variables. Panel (B) and (C) show the accumulated local effect plots (ALE), i.e., the change of the predicted occurrence probability, for the Bioclim variables 3 (Isothermality) and 4 (Temperature Seasonality).

# Conclusion

'cito' is a powerful and versatile R package for building and training fully-connected neural networks with a formula syntax. The package seamlessly integrates into the R regression ecosystem and removes many hurdles in using neural networks for inexperienced users, but also saves programming time for experienced users who just want to build simple neural networks. The unique combination of features provided by 'cito', such as training on a GPU, using custom loss functions, baseline loss, confidence intervals, modern DL training techniques such as continue training, learning rate scheduler or early stopping cannot be found in other packages. Future releases of 'cito' aim to implement additional functionalities such as internal cross validation for hyperparameter optimization, gradient based methods for hyperparameter tuning and the integration of recurrent and convolutional neural networks.

## Acknowledgements

We thank Guillaume Blanchet and two anonymous reviewers for their valuable comments and suggestions.

## Conflict of interest statement

The authors declare that they have no conflicts of interest.

## Data and code availability

The processed datasets for the species distribution model (African elephant) are available from Angelov, 2020. We used version 1.0.2 of the 'cito' package. The 'cito' package can be downloaded from CRAN, the code to reproduce the analysis and the benchmark can be found in the following repository https://github.com/citoverse/Amesoeder-et-al-2023. Upon acceptance of the manuscript, we will produce a persistent Zenodo snapshot of this repository, which also includes a doi. Documentation and rendered vignettes (under articles) can be found on https://citoverse.github.io/cito/ or on the CRAN website of the package at https://cran.r-project.org/web/packages/cito/index.html.